\documentclass[10pt,twocolumn,letterpaper]{article}

\usepackage{cvpr}              

\usepackage{graphicx}
\usepackage{amsmath}
\usepackage{amssymb}
\usepackage{booktabs}
\usepackage{url}

\usepackage{amsfonts}
\usepackage{soul}
\usepackage{tabularx}
\usepackage{color}
\usepackage{multirow}
\usepackage{cite}
\usepackage{bbding}

\usepackage{amssymb}
\usepackage{pdfrender}
\newcommand*{\boldcheckmark}{%
  \textpdfrender{
    TextRenderingMode=FillStroke,
    LineWidth=.5pt, 
  }{\checkmark}%
}

\usepackage[pagebackref,breaklinks,colorlinks]{hyperref}

\usepackage[capitalize]{cleveref}
\crefname{section}{Sec.}{Secs.}
\Crefname{section}{Section}{Sections}
\Crefname{table}{Table}{Tables}
\crefname{table}{Tab.}{Tabs.}

\def\cvprPaperID{3762} 
\def\confName{CVPR}
\def\confYear{2023}

\begin{document}

\title{Perception-Oriented Single Image Super-Resolution using Optimal Objective Estimation}

\author{Seung Ho Park$^{1,2}$, Young Su Moon$^{2}$, Nam Ik Cho$^{1}$\\
$^{1}${Department of Electrical and Computer Engineering, INMC, Seoul National University, Korea}\\
$^{2}${Visual Display Division, Samsung Electronics, Korea}\\
}
\maketitle

\begin{abstract}
Single-image super-resolution (SISR) networks trained with perceptual and adversarial losses provide high-contrast outputs compared to those of networks trained with distortion-oriented losses, such as L1 or L2. However, it has been shown that using a single perceptual loss is insufficient for accurately restoring locally varying diverse shapes in images, often generating undesirable artifacts or unnatural details. For this reason, combinations of various losses, such as perceptual, adversarial, and distortion losses, have been attempted, yet it remains challenging to find optimal combinations. Hence, in this paper, we propose a new SISR framework that applies optimal objectives for each region to generate plausible results in overall areas of high-resolution outputs. Specifically, the framework comprises two models: a predictive model that infers an optimal objective map for a given low-resolution (LR) input and a generative model that applies a target objective map to produce the corresponding SR output. The generative model is trained over our proposed objective trajectory representing a set of essential objectives, which enables the single network to learn various SR results corresponding to combined losses on the trajectory. The predictive model is trained using pairs of LR images and corresponding optimal objective maps searched from the objective trajectory. Experimental results on five benchmarks show that the proposed method outperforms state-of-the-art perception-driven SR methods in LPIPS, DISTS, PSNR, and SSIM metrics. The visual results also demonstrate the superiority of our method in perception-oriented reconstruction.
The code and models are available at \url{https://github.com/seungho-snu/SROOE}.
\end{abstract}
\vspace{-0.1cm}

\section{Introduction}
\label{sec:intro}
The purpose of single image super-resolution (SISR) is to estimate a high-resolution (HR) image corresponding to a given low-resolution (LR) input. SISR has many applications, mainly as a pre-processing step of computer vision or image analysis tasks, such as medical \cite{Park2003SPM, Greenspan2009CJ, You2020TMI}, surveillance \cite{Uiboupin2016SIU, farooq2021human}, and satellite image analysis \cite{Luo2017GRSL, Song2018JSTAEORS}. However, SISR is an ill-posed problem in that infinitely many HR images correspond to a single LR image. 
Recently, the performance of SISR has been greatly improved by adopting deep neural networks \cite{dong2014learning, dong2015image, tai2017memnet, lim2017enhanced, lai2017deep, zhang2018residual, zhang2018image,he2016identity,ICML-2015-IoffeS,KingmaB14}. 
Pixel-wise distortion-oriented losses (L1 and L2) were widely used in early research, which helped to obtain a high signal-to-noise ratio (PSNR). However, these losses lead the model to generate an average of possible HR solutions, which are usually blurry and thus visually not pleasing.

\begin{figure}[t!]
\newcolumntype{Z}
{>{\centering\arraybackslash}X}
\renewcommand{\tabcolsep}{1pt}
\centering
\scriptsize
	\begin{tabularx}{\linewidth}{Z Z Z Z }
	{\includegraphics[width=0.8\linewidth]{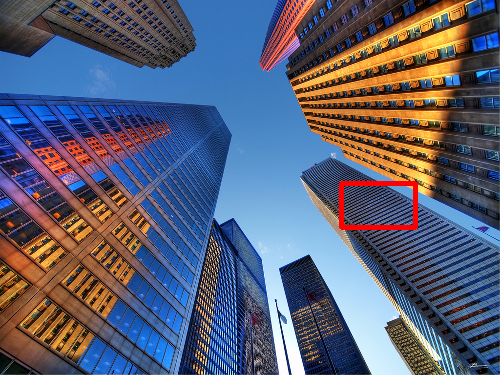}}&	
	{\includegraphics[width=1.0\linewidth]{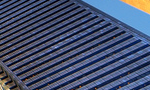}}& 
	{\includegraphics[width=1.0\linewidth]{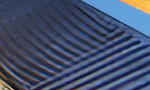}}&
	{\includegraphics[width=1.0\linewidth]{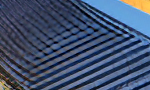}}\\	
	img\_012  & HR &SRGAN\cite{ledig2017photo}& ESRGAN\cite{wang2018esrgan} \\
	(Urban100)&\scriptsize{(PSNR$\uparrow$ / LPIPS$\downarrow$)}& (21.06 / 0.1521) & (20.80 / 0.1182) \\
	{\includegraphics[width=0.8\linewidth]{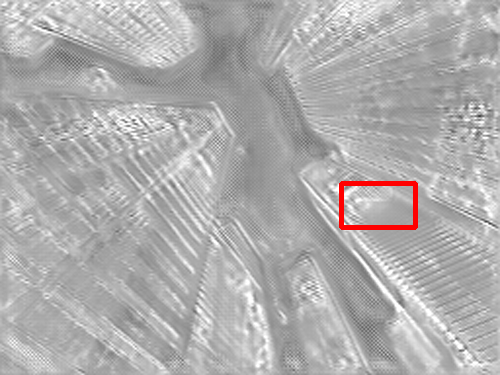}}&
	{\includegraphics[width=1.0\linewidth]{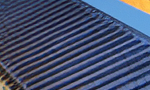}}&
	{\includegraphics[width=1.0\linewidth]{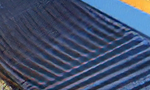}}&
	{\includegraphics[width=1.0\linewidth]{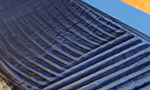}}	\\
	OOE Map & SROOE (Ours)& RankSRGAN\cite{zhang2019ranksrgan}& SPSR\cite{ma2020structure}\\
	& (\textcolor{red}{21.71} / \textcolor{red}{0.1030}) & (21.05 / 0.1462) & (21.27 / 0.1241)  \\
	\end{tabularx}
	\\[1em]
\begin{minipage}[t]{1.0\linewidth}
  \centerline{\includegraphics[width=0.70\linewidth]{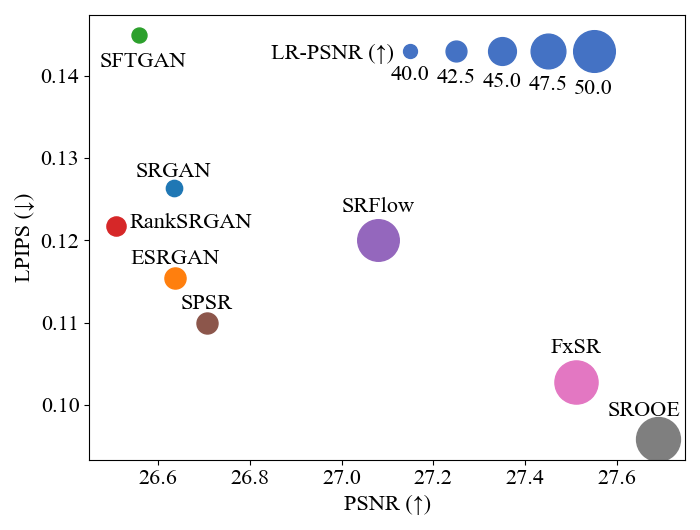}}
\end{minipage}
\vspace{-0.4cm}
\caption{Visual and quantitative comparison. The proposed SROOE shows a higher PSNR, LR-PSNR~\cite{Lugmayr_2021_CVPR} and lower LPIPS~\cite{zhang2018unreasonable} than other state-of-the-art methods, \ie, lower distortion and higher perceptual quality.}
\label{fig:intro}
\vspace{-0.4cm}
\end{figure}

Subsequently, perception-oriented losses, such as perceptual loss~\cite{johnson2016perceptual} and generative adversarial loss~\cite{goodfellow2014generative}, were introduced to overcome this problem and produce realistic images with fine details~\cite{ledig2017photo}. Although these perception-oriented losses are used for various SR methods \cite{wang2018esrgan, sajjadi2017enhancenet, xu2017learning}, they also bring undesirable side effects such as unnatural details and structural distortions. To alleviate these side effects and improve perceptual quality, various methods, such as the ones employing specially designed losses~\cite{soh2019natural, zhang2019ranksrgan} and conditional methods utilizing prior information and additional network branches ~\cite{wang2018recovering, ma2020structure}, have been introduced. Meanwhile, different from the conventional SR methods, which optimize a single objective, some studies tried to apply multiple objectives to generate more accurate HR outputs. 
However, some of them~\cite{wang2018esrgan, wang2019deep, Park2022Access-FxSR} applied image-specific objectives without consideration for the regional characteristics, and the other~\cite{rad2019srobb} used region-specific objectives for the regions obtained using semantic image segmentation with a limited number of pre-defined classes.

In this paper, we propose a new SR framework that finds a locally optimal combination of a set of objectives in the continuous sample space, resulting in regionally optimized HR reconstruction. The upper part of Fig. \ref{fig:intro} shows a visual comparison of our results with those of state-of-the-art perception-oriented methods. We can see that our SR method using optimal objective estimation (OOE), called SROOE, generates more accurate structures. The lower part of Fig.~\ref{fig:intro} shows that the SROOE is located on the far right and bottom, corresponding to the position where both PSNR and LPIPS~\cite{zhang2018unreasonable} are desirable.

For this purpose, our SR framework consists of two models: a predictive model that infers the most appropriate objectives for a given input, and a generative model that applies locally varying objectives to generate the corresponding SR result. The main challenge is to train a single generator to learn continuously varying objectives over the different locations. For this, the objective is defined as the weighted sum of several losses, and we train the generator with various sets of weights. Meanwhile, the role of the predictor is to estimate appropriate weights for a given image input.

For efficient training, we do not learn over the entire objective space spanned by the weight vector, but find a set of several objectives that have high impacts on optimization at each vision level and are close to each other in the objective space. This is because proximity between objectives improves the efficiency of learning and increases the similarity of their results, which helps reduce side effects. In addition, we train the generative model on a set of objectives on our defined trajectory, which is formed by connecting the selected objectives such that the trajectory starts with an objective suitable for a low-vision level and progresses through objectives suitable for higher levels. This enables us to replace high-dimensional weight vector manipulation with simple one-dimensional trajectory tracking, thereby simplifying the training process. The predictive model is trained using a dataset with pairs of LR images and corresponding optimal objective maps. We obtain these optimal training maps by using a grid search on the generator's objective trajectory. 

Regarding the network structure, we employ spatial feature transform (SFT) layers~\cite{wang2018recovering} in the generator to flexibly change the network's behavior according to the objective. Our flexible model trained in this way has three advantages. First, the generalization capability to diversely structured images is improved since the network learns various cases. Second, the SR results are consistent with respect to the trajectory and given input. Third, the high-dimensional weight vector for loss terms can be replaced with a vector function with a one-dimensional input, and thus the optimal loss combinations can be easily found and controlled. 

Our contributions are summarized as follows. (1) We propose an SISR framework that estimates and applies an optimal combination of objectives for each input region and thus produces perceptually accurate SR results. (2) While this approach requires training with various weighted combinations of losses, which needs the search on a high-dimensional weight vector space, we introduce an efficient method for exploring and selecting objectives by defining the objective trajectory controlled by a one-dimensional variable. (3) We propose a method for obtaining optimal objective maps over the trajectory, which are then used to train the objective estimator. (4) Experiments show that our method provides both high PSNR and low LPIPS, which has been considered a trade-off relation.

\section{Related Work}
\textbf{Distortion-oriented SR.}
Dong \etal~\cite{dong2014learning} first proposed a convolutional neural network (CNN)-based SR method that uses a three-layer CNN to learn the mapping from LR to HR. Since then, many deeper CNN-based SISR frameworks have been proposed \cite{kim2016accurate, lim2017enhanced}. Ledig \etal~\cite{ledig2017photo} proposed SRResNet, which uses residual blocks and skip-connections to further enhance SR results. Since Huang \etal~\cite{Huang2017CVPR-DenseNet} proposed DenseNet, the dense connections have become prevalent in SR networks \cite{zhang2018residual, wang2018esrgan, Ahn2018ECCV-CARN, Haris2021TPAMI-DBPN, Anwar2022TPAMI-DRLN}. Zhang \etal~\cite{zhang2018image} introduced RCAN, which employs channel attention and improves the representation ability of the model and SR performance. More recently, SwinIR~\cite{Liang2021ICCVW-SwinIR} and Uformer~\cite{Wang2022CVPR-Uformer} reported excellent SISR performance by using the Swin Transformer architecture \cite{Liu2021ICCV-SwinT} and locally-enhanced window (LeWin) Transformer block, respectively. While there are many architectures for the SR as listed above, we employ plain CNN architectures as our predictor and generator. The structure is not an issue in this paper, and various CNNs and Transformers can be tried instead of our architecture.

\begin{figure*}[ht]
  \centering
\begin{minipage}[t]{1.0\linewidth}
    \centerline{\includegraphics[width=0.85\linewidth]{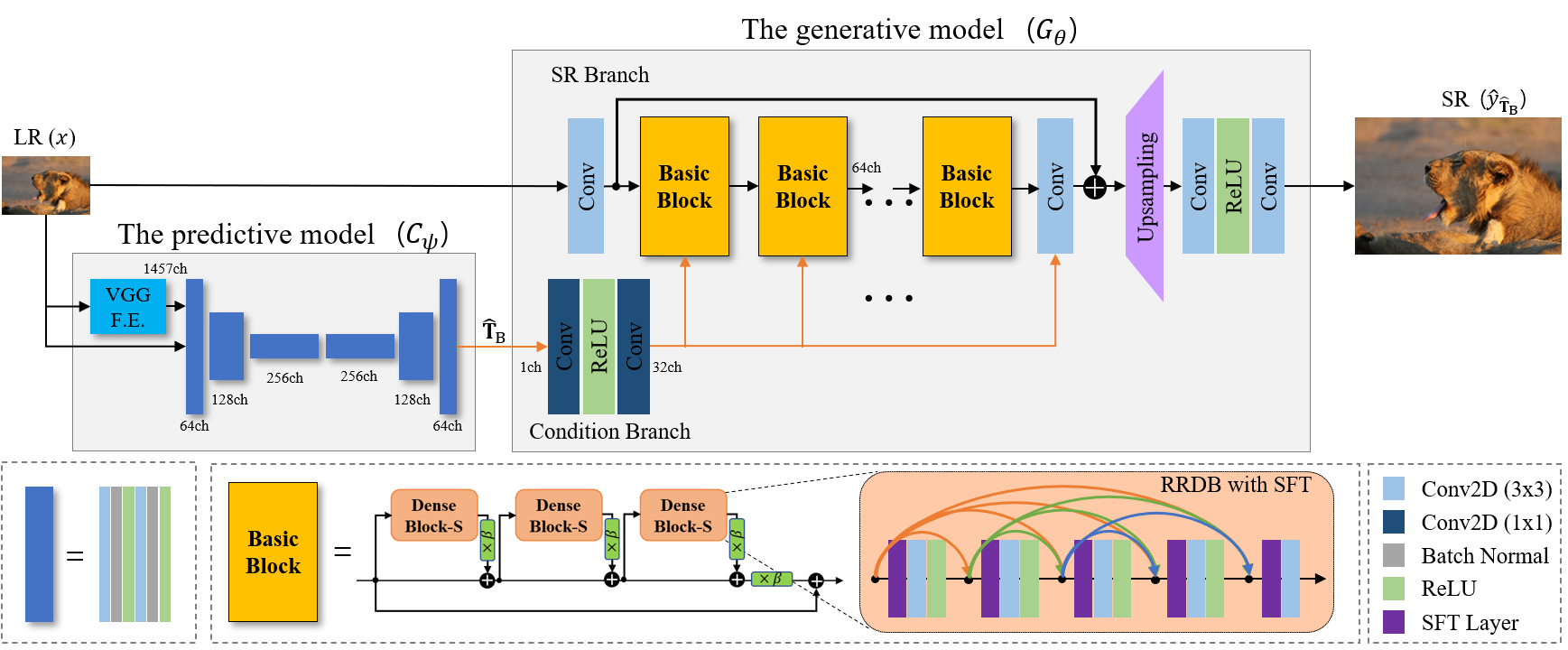}}\vfill
\end{minipage}
\caption{Architecture of the proposed method. The predictive model generates the optimal objective map ${\mathbf{\hat{T}}}_{B}$, which is fed to the generative model. The input LR image is super-resolved through our Basic Blocks and other elements of the generator, which are controlled by the map from the Condition Branch.
}
\label{fig:network-architecture}
\vspace{-0.3cm}
\end{figure*}

\textbf{Perception-oriented SR.}
Because the pixel losses, such as L1 and L2, do not consider perceptual quality, the results of using such losses often lack high-frequency details \cite{wang2004image, Wang2021TPAMI-survey}. Meanwhile, Johnson \etal~\cite{johnson2016perceptual} proposed a perceptual loss to improve the visual quality of the output. Ledig \cite{ledig2017photo} introduced SRGAN utilizing adversarial loss~\cite{goodfellow2014generative}, which can generate photo-realistic HR images. Wang \etal ~\cite{wang2018esrgan} enhanced this framework by introducing Residual-in-Residual Dense Block (RRDB), named ESRGAN.

However, these perception-oriented SR models entail undesirable artifacts, such as unexpected textures on a flat surface. To alleviate such artifacts and/or further improve the perceptual quality, various methods have been proposed. Soh \etal~\cite{soh2019natural} introduced NatSR, where they designed a loss to suppress aliasing. Wang \etal \cite{wang2018recovering} proposed the use of semantic priors for generating semantic-specific details by using SFT layers.
Zhang \etal~\cite{zhang2019ranksrgan} proposed a Ranker that learns the behavior of perceptual metrics.
Ma \etal~\cite{ma2020structure} proposed a structure-preserving super-resolution (SPSR) to alleviate geometric distortions. Liang \etal~\cite{Liang2022CVPR-LDL} proposed locally discriminative learning between GAN-generated artifacts and realistic details.
However, Blau~\cite{blau2018perception} argued that it is difficult to simultaneously achieve perceptual quality enhancement and distortion reduction because they involve a trade-off relationship. In this regard, there was an SR challenge~\cite{Blau2018eccv-Workshops} focused on the trade-off between generation accuracy and perceptual quality. One of the main claims of this paper is that we can further reduce distortion and increase perceptual quality simultaneously, as shown in Fig.~\ref{fig:intro}.

\section{Methods}
\subsection{Proposed SISR Framework}
An overview of our SISR framework is presented in Fig.~\ref{fig:network-architecture}. Our framework consists of a predictive model ${C}_{\psi}$ and generative model $G_\theta$, parameterized by ${\psi}$ and ${\theta}$, respectively.
Model ${C}_{\psi}$ infers an LR-sized optimal objective map $\mathbf{\hat{T}_{B}}$ for a given LR input $x$, and $G_\theta$ applies it to produce the corresponding SR output, which is as similar as possible to its corresponding HR counterpart $y$, as follows:
\begin{equation}
\hat{y}_{{\mathbf{\hat{T}}}_{B}}={G}_{\theta}\left(x|{\mathbf{\hat{T}}}_{B}\right),
\label{eqn:G_thetai}
\end{equation}
\begin{equation}
\hat{\mathbf{T}}_{B}={C}_{\psi}\left(x\right).
\label{eqn:OOE_C_psi}
\end{equation}

\subsection{Proposed Generative Model}
Since using a single fixed objective cannot generate optimized HR results for every image region, it is beneficial to apply regionally different losses regarding the input characteristics. However, training multiple SR models, each of which is trained with a different objective, is impractical because it requires large memory and long training and inference times~\cite{dosovitskiy2019you}. Hence, in this paper, we propose a method to train a single SR model that can consider locally different objectives.

\textbf{Effective Objective Set.}
We first investigate which objectives need to be learned for accurate SR. For perception-oriented SR \cite{ledig2017photo, sajjadi2017enhancenet}, the objective is usually a weighted sum of pixel-wise reconstruction loss $\mathcal{L}_{rec}$, adversarial loss $\mathcal{L}_{adv}$, and perceptual loss $\mathcal{L}_{per}$, as follows:
\begin{equation}
\mathcal{L}={\lambda}_{rec}\cdot\mathcal{L}_{rec}+{\lambda}_{adv}\cdot\mathcal{L}_{adv}+\sum_{per_l}\lambda_{per_l}\cdot{L}_{per_l}
\label{eqn:loss-function},
\end{equation}
\begin{equation}
{L}_{per_l}=\mathbb{E}\left[\lVert{\phi_{per_l}(\hat{y})-\phi_{per_l}(y))}\rVert_{1}\right],
\label{eqn:per-loss-function}
\end{equation}
\begin{equation}
{per_l}\in\{\text{V12, V22, V34, V44, V54}\},
\label{eqn:vggset}
\end{equation}
where ${\lambda}_{rec}$,  ${\lambda}_{adv}$, and ${\lambda}_{per_l}$ are weighting parameters for the corresponding losses, and ${\phi}_{per_l}(\cdot)$ represents feature maps of the input extracted at layer $per_l$ of the 19-layer VGG network, where five layers denoted in Eqn.~\ref{eqn:vggset} are considered as in \cite{SimonyanZ14a, rad2019srobb,Park2022Access-FxSR}. Since the receptive field becomes larger as we progress deeper into the VGG network~\cite{SimonyanZ14a}, features of shallow layers such as V12 and V22 and deeper layers such as V34, V44, and V54 correspond to relatively low-level and higher-level vision, respectively~\cite{rad2019srobb}. 

\newcolumntype{M}[1]{>{\centering\arraybackslash}m{#1}}
\begin{table}[!t]
\caption{Performance comparison of SR results of ESRGAN models with different weight vectors for perceptual loss. 
Among the objectives in Sets $A$ and $B$, except for $\boldsymbol{\lambda_{0}}$, the 1st and the 2nd best performances for each column are highlighted in \textcolor{red}{red} and \textcolor{blue}{blue}.}
\vspace{-0.4cm}
\begin{center}
\scriptsize
\renewcommand{\tabcolsep}{1pt}
\begin{tabular}{M{2.9mm}|M{5.5mm}|M{5mm}M{5mm}M{5mm}M{5mm}M{5.2mm}||M{5.1mm}|M{5.1mm}|M{5.1mm}|M{5.1mm}|M{5.1mm}||M{5.8mm}|M{6mm}}
\toprule
 \multirow{2}*{Set}& {objec} &\multicolumn{5}{c||}{$\boldsymbol{\lambda_{per}}$}&\multicolumn{5}{c||}{Normalized $L_{per_l}$}&\multicolumn{2}{c}{Metric}\\
 & -tive &[$\lambda_{\text{V12}}$,&$\lambda_{\text{V22}}$,&$\lambda_{\text{V34}}$,&$\lambda_{\text{V44}}$,&$\lambda_{\text{V54}}$]&${L}_{\text{V12}}$&${L}_{\text{V22}}$&${L}_{\text{V34}}$&${L}_{\text{V44}}$&${L}_{\text{V54}}$&PSNR&LPIPS\\
\midrule
  & $\boldsymbol{\lambda_{0}}$&[0.0, 	& 0.0, &0.0, 	&0.0, &0.0] &0.00	&0.00	&1.00	&1.00	&1.00&25.48&0.1960\\
\midrule
$A$& $\boldsymbol{\lambda_{1}}$&[1.0, 	& 0.0, & 0.0, 	&0.0, &0.0] & \textcolor{blue}{0.71}	& 0.53	& 0.76	& 0.38	& 0.25&\textcolor{blue}{23.95}&{0.1124}\\
  & $\boldsymbol{\lambda_{2}}$&[0.0, & 1.0, & 0.0, 	&0.0, &0.0] &0.72	&\textcolor{blue}{0.26}	&0.35	&0.22	&0.15&23.84&0.1125\\
  & $\boldsymbol{\lambda_{3}}$&[0.0, 	&0.0, &1.0,	   	&0.0, &0.0] &0.81	&0.51	&\textcolor{red}{0.00}	&\textcolor{red}{0.02}	&0.04&23.66&{0.1124}\\
  & $\boldsymbol{\lambda_{4}}$&[0.0, &0.0, &0.0,  	&1.0,	&0.0] &0.92	&0.78	&0.51	&0.11	&0.05&23.28&0.1158\\
  & $\boldsymbol{\lambda_{5}}$&[0.0, & 0.0,& 0.0, 	&0.0, &1.0]	&1.00	&1.00	&0.93	&0.32	&0.12&23.00&0.1232\\
\midrule
$B$& $\boldsymbol{\lambda_{1}}$&[1.0,& 0.0,& 0.0,&0.0,&0.0]& \textcolor{blue}{0.71}	& 0.53	& 0.76	& 0.38	& 0.25&\textcolor{blue}{23.95}&0.1124\\
  & $\boldsymbol{\lambda_{1\text{-}2}}$&[1/2,&1/2,&0.0,&0.0,&0.0]&\textcolor{red}{0.64}	&\textcolor{red}{0.23}	&0.34	&0.20	&0.13&\textcolor{red}{24.08}&\textcolor{red}{0.1075}\\
  & $\boldsymbol{\lambda_{1\text{-}3}}$&[1/3,&1/3,&1/3,&0.0,&0.0]&0.78	&0.42	&\textcolor{blue}{0.10}	&0.04	&\textcolor{blue}{0.03}&23.81&0.1112\\
  & $\boldsymbol{\lambda_{1\text{-}4}}$&[1/4,&1/4,&1/4,&1/4,&0.0]&0.78	&0.46	&0.18	&\textcolor{red}{0.02}	&\textcolor{red}{0.01}&23.68&\textcolor{blue}{0.1110}\\
\midrule
\midrule
\multicolumn{12}{c||}{$\text{ESRGAN-OOS}_{A}$}&{24.03}&{0.0848}\\
\multicolumn{12}{c||}{$\text{ESRGAN-OOS}_{B}$}&{24.21}&{0.0848}\\
\bottomrule
\end{tabular}
\end{center}
\label{tab:two-objective-set}
\vspace{-0.3cm}
\end{table}

\begin{figure}[!t]
\setlength{\arrayrulewidth}{1.0pt}
\newcolumntype{Z}{>{\centering\arraybackslash}X}
\renewcommand{\tabcolsep}{1pt}
\centering
\footnotesize
	\begin{tabularx}{\linewidth}{Z Z}
	{\includegraphics[width=0.70\linewidth]{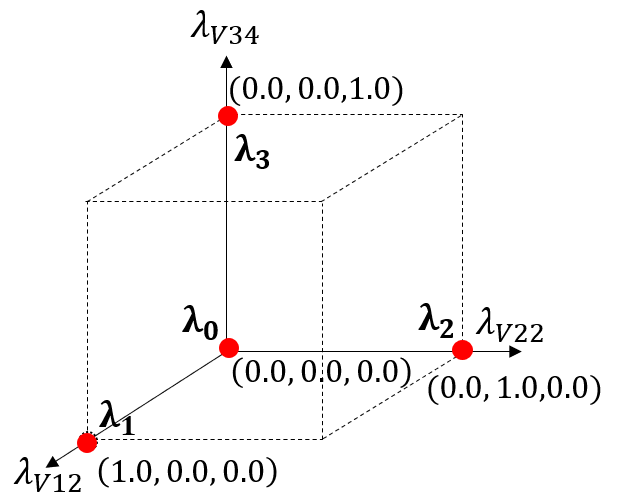}}	&
	{\includegraphics[width=0.70\linewidth]{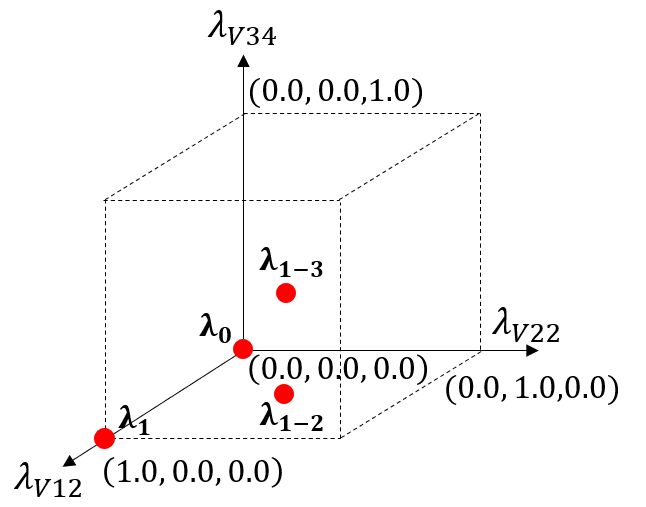}}	\\
	\end{tabularx}
\caption{Set $A$ (left) and set $B$ (right) in the objective space.}
\label{fig:objectives-in-the-spacet}
\vspace{-0.3cm}
\end{figure}

To find an effective set of objectives, we define an SR objective space. Since the objective for SR is a weighted sum of seven loss terms, as in Eqn. \ref{eqn:loss-function}, an objective space is spanned by these basis loss terms, and any objective can be expressed by a seven-dimensional vector of weighting parameters, $\boldsymbol{\lambda}\in\mathbb{R}^{7}$ as 
\begin{equation}
\boldsymbol{\lambda}=[\lambda_{rec}, \lambda_{adv}, \boldsymbol{\lambda_{per}}],
\label{eqn:lambda}
\end{equation}
where $\boldsymbol{\lambda_{per}}\in\mathbb{R}^{5}$ is a weight vector for perceptual loss.

Table~\ref{tab:two-objective-set} compares two objective sets, $A$ and $B$, defined as shown in Fig.~\ref{fig:objectives-in-the-spacet}.
Because ESRGAN~\cite{wang2018esrgan} is the base model for this comparison, for all objectives in the table, except for $\boldsymbol{\lambda_{0}}$, $\lambda_{rec}$ and $\lambda_{adv}$ are set to ${1}\times{10}^{-2}$ and ${5}\times{10}^{-3}$, respectively. These are the same as those for ESRGAN, except that $\boldsymbol{\lambda_{per}}$ changes, where $\lVert{\boldsymbol{\lambda_{per}}}\rVert_{1}=1$.
In particular, in terms of $\boldsymbol{\lambda_{per}}$, whereas each objective in set $A$ has weights for only one of the five VGG feature spaces, each objective in set $B$ has equal weights for each loss in the feature space lower than the target vision level.
Therefore, an objective corresponding to a high vision level also includes the losses for the lower-level feature spaces.
Meanwhile, because $\boldsymbol{\lambda_{0}}$ corresponds to a distortion-oriented RRDB model~\cite{wang2018esrgan}, its $\lambda_{rec}$ and $\lambda_{adv}$ are set to ${1}\times{10}^{-2}$ and 0, respectively. Note that $\boldsymbol{\lambda_{0}}$ is included in both sets $A$ and $B$.

\begin{figure}[!t]
\newcolumntype{Z}{>{\centering\arraybackslash}X}
\renewcommand{\tabcolsep}{1pt}
\centering
\scriptsize
	\begin{tabularx}{\linewidth}{Z Z Z Z}
	{\includegraphics[width=0.95\linewidth]{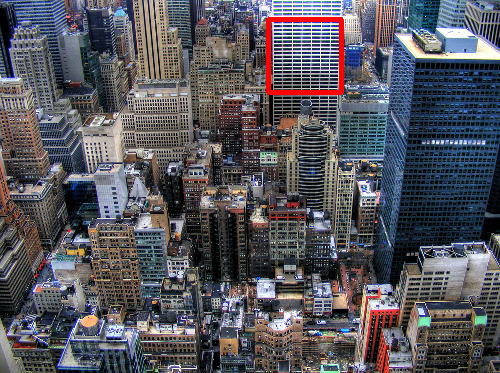}}	&
	{\includegraphics[width=0.95\linewidth]{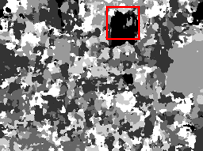}}	&
	{\includegraphics[width=0.95\linewidth]{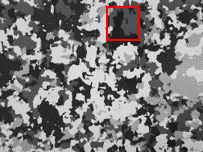}}	&
	{\includegraphics[width=0.75\linewidth]{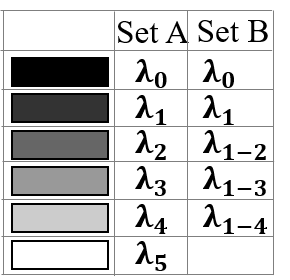}}\\
	img\_073 & $\text{OOS}_{A}$ &$\text{OOS}_{B}$ & intensity of each $\boldsymbol{\lambda}$\\
	\end{tabularx}
	\begin{tabularx}{\linewidth}{Z | Z Z | Z Z}	
	{\includegraphics[width=0.9\linewidth]{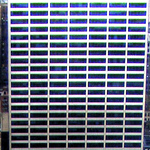}}&
	{\includegraphics[width=0.9\linewidth]{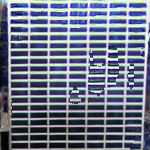}}&
	{\includegraphics[width=0.9\linewidth]{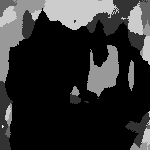}}&
	{\includegraphics[width=0.9\linewidth]{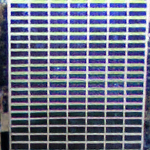}}&
	{\includegraphics[width=0.9\linewidth]{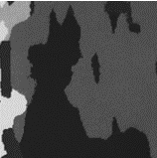}}\\
	 {HR}  & \multicolumn{2}{c|}{$\text{ESRGAN-OOS}_{A}$ / $\text{OOS}_{A}$ } & \multicolumn{2}{c}{$\text{ESRGAN-OOS}_{B}$ / $\text{OOS}_{B}$} \\	
	\end{tabularx}
\vspace{-0.1cm}
\caption{ESRGAN-OOS and OOS using Sets $A$ and $B$.}
\label{fig:VGG_Toos}
\vspace{-0.3cm}
\end{figure}

In Table~\ref{tab:two-objective-set}, the normalized versions (min-max feature scaling) of the averaged $L_{per_l}$ from Eqn.~\ref{eqn:per-loss-function} for five datasets (BSD100~\cite{Martin2001ICCV-BSD100}, General100~\cite{dong2016accelerating}, Urban100~\cite{huang2015single}, Manga109~\cite{matsui2017sketch}, and DIV2K~\cite{agustsson2017ntire}) are reported.
For all feature spaces, including the targeted V12 and V22 feature spaces, $\boldsymbol{\lambda_{1\text{-}2}}$ in set $B$ has smaller L1 errors than those of $\boldsymbol{\lambda_{1}}$ and $\boldsymbol{\lambda_{2}}$ in set $A$. Moreover, $\boldsymbol{\lambda_{1\text{-}4}}$ exhibits smaller errors than those of $\boldsymbol{\lambda_{4}}$ and $\boldsymbol{\lambda_{5}}$.
Although $\boldsymbol{\lambda_{1\text{-}3}}$ has slightly more errors in the V34 feature space than that of $\boldsymbol{\lambda_{3}}$, it has less errors in the V12 and V22 feature spaces therefore, $\boldsymbol{\lambda_{1\text{-}3}}$ has relatively less distortion than $\boldsymbol{\lambda_{3}}$ overfitted to the V34 feature space.
That is supported by the fact that most of the objectives in set $B$, including $\boldsymbol{\lambda_{1\text{-}3}}$, have better PSNR and LPIPS on Urban100~\cite{huang2015single} than those in set $A$.

To examine the SR result with locally appropriate objectives applied using set $A$, we mix the six SR results of $\text{ESRGAN-}\boldsymbol{\lambda_a}$, where $\boldsymbol{\lambda_a}\in{A}$, by selecting the SR result with the lowest LPIPS for each pixel position, as follows:
\begin{equation}
    {{y}^\ast_{A}(i,j)}={\hat{y}_{{\mathbf{T}^\ast_{A}(i,j)}}(i,j)},
\end{equation}
\begin{equation}
    {\mathbf{T}^\ast_{A}(i,j)}=\arg\min_{\boldsymbol{\lambda_{a}}\in{A}}{\mathbf{LPIPS}_{\boldsymbol{\lambda_a}}(i,j)},
\end{equation}
\begin{equation}
    \mathbf{LPIPS}_{\boldsymbol{\lambda_a}}=LPIPS\left(y,\hat{y}_{\boldsymbol{\lambda_a}}\right),
\end{equation}
where $\hat{y}_{\boldsymbol{\lambda_a}}$ is the SR result of $\text{ESRGAN-}\boldsymbol{\lambda_a}$. The $LPIPS$ function computes the perceptual distance between two image patches for each pixel position, producing an LPIPS map, $\mathbf{LPIPS}$, of the input image size~\cite{Lugmayr_2021_CVPR, Park2022Access-FxSR}. 
The LPIPS metric in Table~\ref{tab:two-objective-set} is the average of this map.
Since $\mathbf{T}^\ast_{A}$ is the optimal objective selection (OOS), $\mathbf{T}^\ast_{A}$ and its SR model for mixing are denoted as $\text{OOS}_{A}$ and $\text{ESRGAN-OOS}_{A}$, respectively. 
The upper part of Fig.~\ref{fig:VGG_Toos} shows an example of $\text{OOS}_{A}$ and $\text{OOS}_{B}$ based on set $A$ and $B$. PSNR and LPIPS~\cite{zhang2018unreasonable} of $\text{ESRGAN-OOS}_{A}$ and $\text{ESRGAN-OOS}_{B}$ are reported in Table \ref{tab:two-objective-set}, where $\text{ESRGAN-OOS}_{B}$ is superior to any single objective model, demonstrating the potential for performance improvement of the locally suitable objective application. 
The lower part of Fig.~\ref{fig:VGG_Toos} shows the side effects caused by mixing the SR results for set $A$ with lower proximities between objectives than those in set $B$, as shown in Fig.~\ref{fig:objectives-in-the-spacet}.
Since $\text{ESRGAN-OOS}_{B}$ in Fig.~\ref{fig:VGG_Toos} has less artifact and better PSNR than those of $\text{ESRGAN-OOS}_{A}$, the proposed set $B$ is more suitable for applying locally appropriate objectives.

\begin{figure}[!t]
\setlength{\arrayrulewidth}{1.0pt}
\newcolumntype{Z}{>{\centering\arraybackslash}X}
\renewcommand{\tabcolsep}{1pt}
\centering
\footnotesize
	\begin{tabularx}{\linewidth}{Z Z}
	{\includegraphics[width=0.85\linewidth]{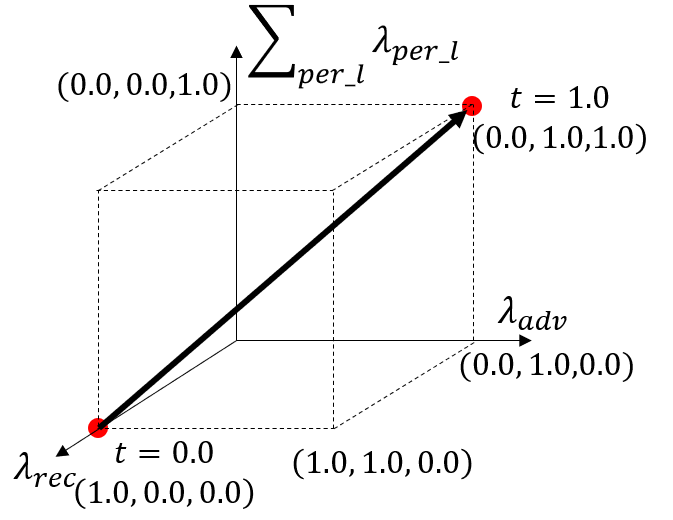}} &	
	{\includegraphics[width=0.85\linewidth]{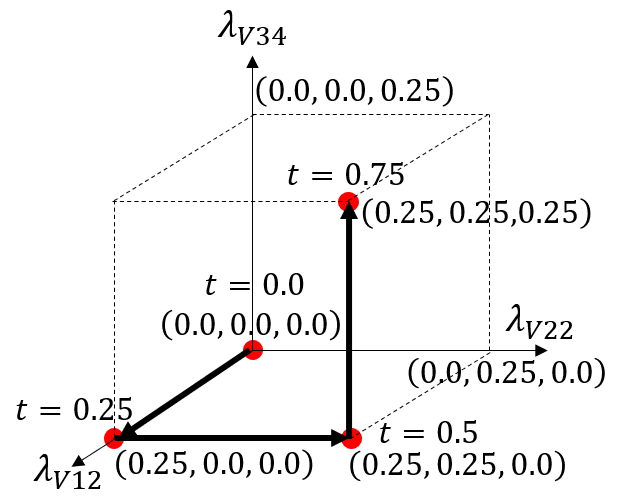}} \\
	(a)  & (b)  \\
	{\includegraphics[width=0.80\linewidth]{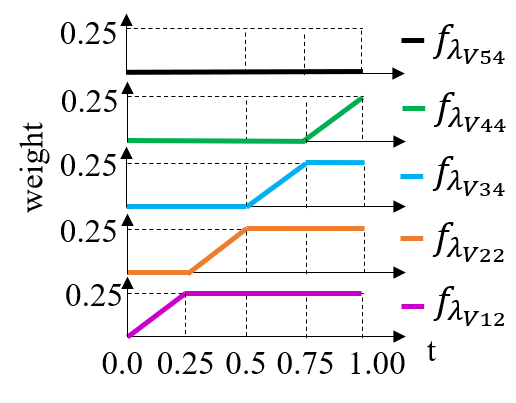}} &
	{\includegraphics[width=0.85\linewidth]{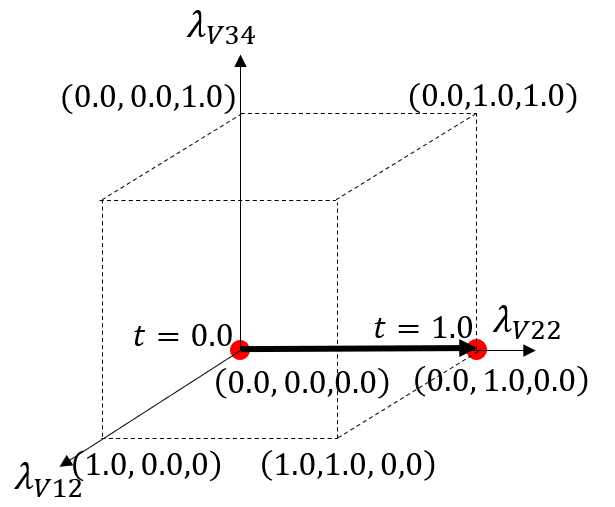}}	\\
	(c)  & (d)  \\
	\end{tabularx}
\vspace{-0.3cm}
\caption{Vector functions for loss weights. ($\boldsymbol{\alpha}$=$\boldsymbol{1}$ and $\boldsymbol{\beta}$=$\boldsymbol{0}$)}
\label{fig:vector-function}
\vspace{-0.3cm}
\end{figure}

\textbf{Learning Objective Trajectory.}
We train our generative model on a set of objectives over the trajectory rather than a single objective, $\boldsymbol{\lambda}$, in Eqn~\ref{eqn:lambda}. The objective trajectory is formed by connecting the selected objectives, i.e. the five objectives of set $B$, starting with an objective for a low-vision level and progressing through objectives for higher levels, i.e. from $\boldsymbol{\lambda_{0}}$ to $\boldsymbol{\lambda_{1\text{-}4}}$. It is parameterized by a single variable $t$
, $\boldsymbol{\lambda}(t)=\left \langle {{\lambda_{rec}}(t), {\lambda_{adv}}(t), \boldsymbol{{\lambda_{per}}}(t)}  \right\rangle$, as follow:
\begin{equation}
\boldsymbol{\lambda}(t)=\boldsymbol{\alpha}{\cdot}\boldsymbol{f}_{\lambda}(t)+\boldsymbol{\beta}, 
\end{equation}
\begin{equation}
\boldsymbol{f}_{\lambda}(t)=\left \langle {f_{\lambda_{rec}}(t), f_{\lambda_{adv}}(t), \boldsymbol{f_{\lambda_{per}}}(t)}  \right\rangle,
\label{eqn:f-lambda-t}
\end{equation}
where ${\boldsymbol{f_{\lambda_{per}}}}(t)\in{\mathbb{R}^{5}}$, $f_{\lambda_{rec}}(t)$, $f_{\lambda_{adv}}(t)$ are weighting functions, $\boldsymbol{\alpha}$ and $\boldsymbol{\beta}$ are the scaling and offset vectors. 
As $\boldsymbol{f}_{\lambda}:\mathbb{R}\rightarrow\mathbb{R}^{7}$, this vector function enables the replacement of high-dimensional weight-vector manipulation with one-dimensional tracking, simplifying the training process.

Specifically, the trajectory design is based on the observation in Table \ref{tab:two-objective-set} that the distortion-oriented RRDB model using $\boldsymbol{\lambda_{0}}$ has smaller L1 errors than those of all ESRGAN models for low-level feature spaces, such as V12 and V22, whereas ESRGAN models have smaller L1 errors for higher-level feature spaces, such as V34, V44, and V54. Thus, we design the weight functions $f_{\lambda_{rec}}$, $f_{\lambda_{adv}}$ and ${\boldsymbol{f_{\lambda_{per}}}}$ such that when $t$ approaches $0$, $f_{\lambda_{rec}}$ increases and \{$f_{\lambda_{adv}},\sum_{per_l}f_{\lambda_{per_l}}$\} decrease to go to $\boldsymbol{\lambda_{0}}$, and conversely to go to $\boldsymbol{\lambda_{1\text{-}4}}$ when $t$ increases to 1, as shown in Fig.~\ref{fig:vector-function}(a), where $\boldsymbol{\alpha}$=$\boldsymbol{1}$, $\boldsymbol{\beta}$=$\boldsymbol{0}$.

In relation to the change in $\sum_{per_l}f_{\lambda_{per_l}}(t)$, we design each of five component functions, $f_{\lambda_{per_l}}(t)$ of ${\boldsymbol{f_{\lambda_{per}}}}(t)$, as shown in Fig.~\ref{fig:vector-function}(c), to obtain the objective trajectory from $\boldsymbol{\lambda_{0}}$ to $\boldsymbol{\lambda_{1\text{-}4}}$ of set $B$ as shown in Fig.~\ref{fig:vector-function}(b), illustrating only three out of five components because of the limitations of 3-dimensional visualization. Thus, as we progress through the trajectory by increasing $t$ from $0$ to $1$, the weighting parameters for the objective start with the distortion-oriented objective, $\boldsymbol{\lambda_{0}}$, and then the losses of higher-vision-level feature spaces and adversarial loss are progressively added, making slight transitions on the objective toward $\boldsymbol{\lambda_{1\text{-}4}}$. Fig.~\ref{fig:vector-function}(d) shows the objective trajectory used for FxSR~\cite{Park2022Access-FxSR}, which uses only the V22 feature space, limiting the performance of the perceptually accurate restoration. 

The proposed objective trajectory can efficiently improve the accuracy and consistency of the SR results. First, we can use any objective on the continuous trajectory from low to high-level vision, which allows the application of more accurate objectives to each region. Second, with regard to consistency, high-level objectives on our proposed trajectory include both low-level and high-level losses, thus also accounting for the low-level objectives.
This weighting method allows the sharing of the structural components reconstructed mainly by low-vision-level objectives between all SR results on the trajectory.
Finally, we need to train a single SR model only once, reducing the number of models required to produce diverse HR outputs~\cite{dosovitskiy2019you, Park2022Access-FxSR}.

\begin{figure}[t!]
\setlength{\arrayrulewidth}{1.0pt}
\centering
\footnotesize
	\begin{tabular}{M{17mm} M{62mm}}
	{\includegraphics[width=1.0\linewidth]{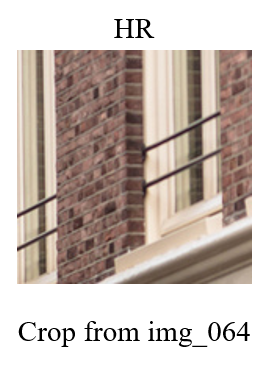}}	&
	{\includegraphics[width=1.0\linewidth]{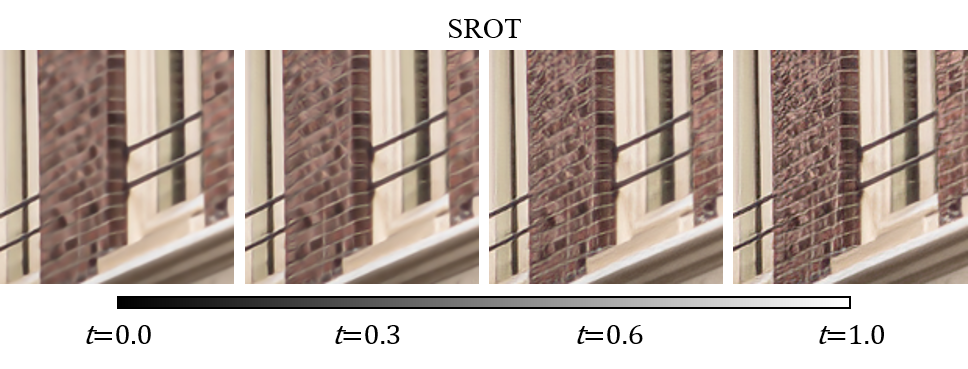}}	\\
	\end{tabular}
\vspace{-0.3cm}
\caption{Changes in the SR results of SROT according to $t$-value.}
\label{fig:SROT-t}
\vspace{-0.3cm}
\end{figure}

\begin{figure}[!t]
\setlength{\arrayrulewidth}{1.0pt}
\newcolumntype{Z}{>{\centering\arraybackslash}X}
\renewcommand{\tabcolsep}{1pt}
\centering
\footnotesize
	\begin{tabularx}{\linewidth}{Z Z}
	{\includegraphics[width=0.85\linewidth]{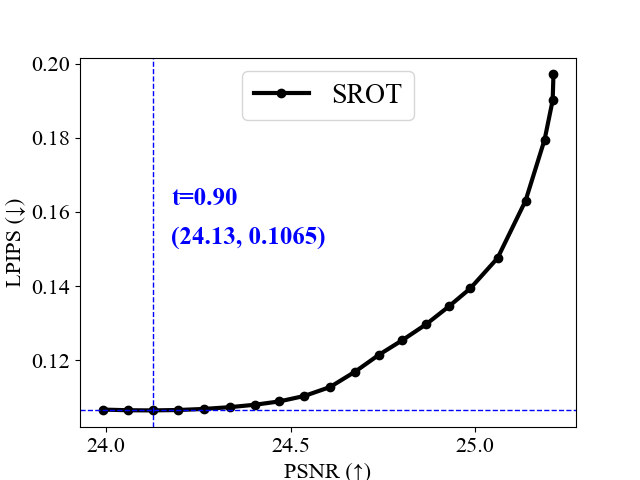}}	&
	{\includegraphics[width=0.85\linewidth]{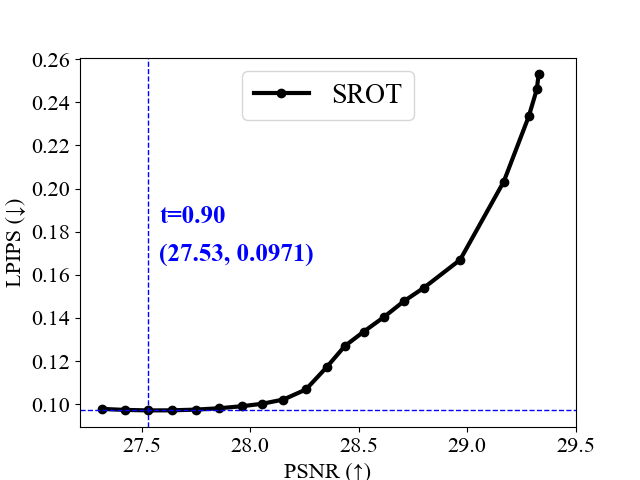}}	\\
	(a)  Urban100 & (b) DIV2K \\
	\end{tabularx}
\caption{Changes in PSNR and LPIPS according to $t$-value.}
\label{fig:PD-curve-t}
\vspace{-0.5cm}
\end{figure}

Fig.~\ref{fig:SROT-t} shows the changes in the SR result of the generative model trained on the objective trajectory (OT) in Fig.~\ref{fig:vector-function}(b), called SROT, as $t$ changes from 0 to 1. Fig.~\ref{fig:PD-curve-t} shows the trade-off curves in the perception-distortion plane according to the change of $t$, where $t$ increases by $0.05$ from $0.0$ to $1.0$ and has 21 sample points.
Each SR result on the curve is obtained by inputting $\mathbf{T}$ with the same $t$ throughout the image, as $\mathbf{T}_t=\mathbf{1}\times{t}$, into the condition branch of the generative model, as follows:
\begin{equation}
\hat{y}_{{\mathbf{T}}_{t}}={G}_{\theta}\left(x|{\mathbf{T}}_{t}\right).
\label{eqn:SR-Tt}
\end{equation}
The horizontal and vertical dotted lines in Fig.~\ref{fig:PD-curve-t} indicate the lowest LPIPS values of the model and the corresponding PSNR values, respectively. The $t$ values at that time are written next to the vertical lines. However, applying a specific $t$ to the entire image still limits SR performance, and optimal $t$ depending on images is unknown at inference time. We take this one step further and present later how to estimate and apply locally optimal objectives.

\textbf{Network Architecture and Training.}
The outline of the generator network is adopted from \cite{Park2022Access-FxSR}, \ie, ${G}_{\theta}$ consists of two streams, an SR branch with 23 basic blocks and a condition branch as shown in Fig.~\ref{fig:network-architecture}. The condition branch takes an LR-sized target objective map $\mathbf{T}$ and produces shared intermediate conditions that can be transferred to all the SFT layers in the SR branch. Since the SFT layers~\cite{wang2018recovering} modulate feature maps by applying affine transformation, they learn a mapping function that outputs a modulation parameter based on $\mathbf{T}$. 
Specifically, $\mathbf{T}_t$, with $t$ randomly changing in the pre-defined range, is fed into the condition branch during training, and this modulation layer allows the SR branch to optimize the changing objective by $t$.
As a result, ${G}_{\theta}$ learns all the objectives on the trajectory and generates SR results with spatially different objectives according to the map at inference time. 
${G}_{\theta}$ is optimized on the training samples $\mathcal{Z}=\left(x,y\right)$ with the distribution $P_{\mathcal{Z}}$, as follows:
\begin{equation}
{\theta}^\ast=\arg\min_{\theta}\mathbb{E}_{\mathcal{Z}\sim\mathcal{P}_{\mathcal{Z}}}\left[\mathcal{L}\left(\hat{y}_{{\mathbf{T}}_{t}},y|t\right)\right],
\label{eqn:optTheta}
\end{equation}
\begin{equation}
\mathcal{L}\left(t\right)={\lambda}_{rec}\left(t\right)\cdot\mathcal{L}_{rec}+{\lambda}_{adv}\left(t\right)\cdot\mathcal{L}_{adv}+\sum_{per_l}\lambda_{per_l}\left(t\right)\cdot{L}_{per_l}.
\label{eqn:loss-function-t}
\end{equation}

\begin{figure}[t!]
\setlength{\arrayrulewidth}{1.0pt}
\newcolumntype{Z}{>{\centering\arraybackslash}X}
\renewcommand{\tabcolsep}{1pt}
\centering
\footnotesize
	\begin{tabularx}{\linewidth}{Z Z Z}
	{\includegraphics[width=0.95\linewidth]{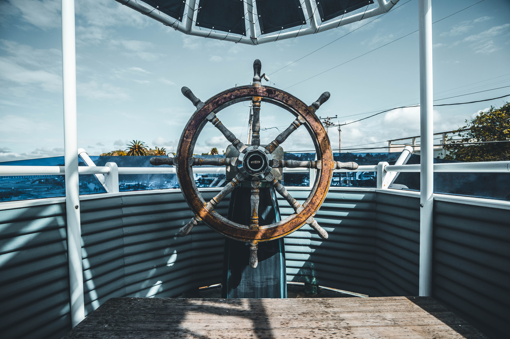}}	&
	{\includegraphics[width=0.95\linewidth]{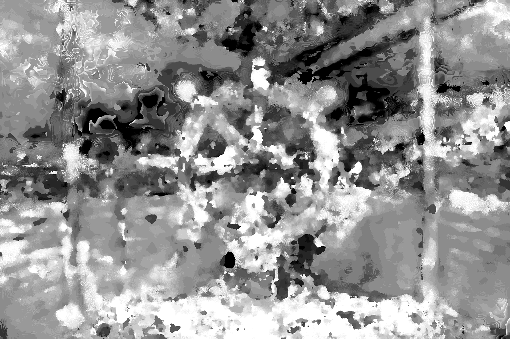}}&
	{\includegraphics[width=0.95\linewidth]{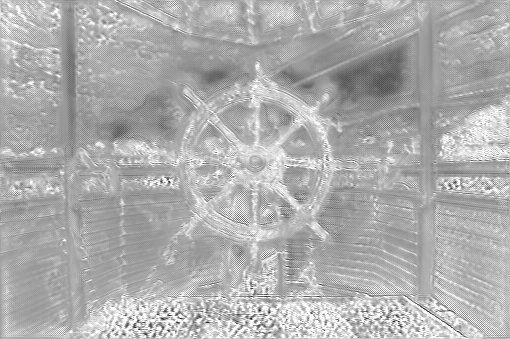}}\\
	image (0824) & OOS $\mathbf{T}^\ast_{S}$ & OOE $\hat{\mathbf{T}}_{B}$\\
	\end{tabularx}
\vspace{-0.3cm}
\caption{The input image, the optimal objective selection $\mathbf{T}^\ast_{S}$ obtained by parameter sweeping, and $\hat{\mathbf{T}}_{B}$ estimated by ${C}_{\psi}$.}
\label{fig:Toos-Tooe}
\vspace{-0.3cm}
\end{figure}

\subsection{Optimal Objective Estimation (OOE)}
To estimate an optimal combination of objectives for each region, we train a predictive model, ${C}_{\psi}$. This model produces an optimal objective map $\hat{\mathbf{T}}_{B}$ estimated for a given LR image, which is then delivered to the generative model in Eqn.~\ref{eqn:G_thetai}.
Since it is hard to find the ground truth map for ${C}_{\psi}$ training, we obtain its approximation $\mathbf{T}^\ast_{S}$ via a simple exhaustive searching to narrow down the range of the best possible values. 
Specifically, we generate a set of 21 SR results by changing $t$ from $0$ to $1$ with a step of $0.05$, and the optimal objective maps are generated by selecting the $t$ with the lowest LPIPS among them for each pixel, as
\begin{equation}
    {\mathbf{T}^\ast_{S}(i,j)}=\arg\min_{{t}\in{S}}{\mathbf{LPIPS}_{t}(i,j)},
\end{equation}
\begin{equation}
    \mathbf{LPIPS}_{t}=LPIPS\left(y,\hat{y}_{\mathbf{T}_{t}}\right),
\end{equation}
where $\mathbf{T}_{t}=\mathbf{1}\times{t},{t}\in{S}=\left\{0.0,0.05,0.10,...,1.0\right\}$.
Fig. \ref{fig:Toos-Tooe} shows an example of the optimal objective selection (OOS) ${\mathbf{T}^\ast_{S}}$, and the SR result using ${\mathbf{T}^\ast_{S}}$, SROOS, can be an upper-bound approximation for the performance of ${G}_{\theta}$ as
\begin{equation}
\hat{y}_{{\mathbf{T}^\ast_{S}}}={G}_{\theta}\left(x|{\mathbf{T}^\ast_{S}}\right).
\label{eqn:SROOS}
\vspace{-0.0cm}
\end{equation}

\newcolumntype{M}[1]{>{\centering\arraybackslash}m{#1}}
\begin{table*}[!t]
\caption{Comparison with state-of-the-art SR methods on benchmarks. 
The 1st and the 2nd best performances for each group are highlighted in \textcolor{red}{red} and \textcolor{blue}{blue}, respectively. LR-PSNR values greater than 45dB are underlined.}
\vspace{-0.4cm}
\begin{center}
\scriptsize
\renewcommand{\tabcolsep}{1pt}
\begin{tabular}{M{12mm}|M{13mm}|M{12mm}M{11mm}|M{11mm}M{12mm}M{12mm}M{11mm}M{11mm}M{11mm}M{11mm}M{11mm}M{11mm}|M{11mm}}
\toprule
     \multicolumn{2}{c|}{}  & \multicolumn{2}{c|}{Distortion-oriented SR}  & \multicolumn{10}{c}{Perception-oriented SR} \\
\midrule
    \multicolumn{2}{c|}{Model}  & { RRDB }\cite{wang2018esrgan}  & SROOE ($\mathbf{T}=\mathbf{0}$) & SRGAN\cite{ledig2017photo} & {ESRGAN}\cite{wang2018esrgan} & {SFTGAN}\cite{wang2018recovering} & RankSR GAN\cite{zhang2019ranksrgan} & SRFlow\cite{lugmayr2020srflow} & {SPSR}\cite{ma2020structure} & FxSR \cite{Park2022Access-FxSR} & SROOE ($\hat{\mathbf{T}}_{B}$) & SROOE ($\hat{\mathbf{T}}_{B}$) & SROOS ($\mathbf{T}^\ast_{S}$) \\ 
\midrule
    \multicolumn{2}{c|}{Training dataset}  & DIV2K & DF2K & DIV2K & \tiny{DF2K+OST} & \tiny{ImageNet+OST}& DIV2K & DF2K & DIV2K & DIV2K& DIV2K & DF2K & DF2K \\ 
\midrule
\midrule
\multirow{5}*{BSD100} 
	& PSNR$\uparrow$ 		& \textcolor{red}{26.53} &\textcolor{blue}{26.45} & 24.13	 & 23.95 & 24.09	 & 24.09 & 24.66 & 24.16 
						& 24.77 & \textcolor{blue}{24.78} &\textcolor{red}{24.87} & {25.07}\\

	& SSIM$\uparrow$ 		& \textcolor{red}{0.7438} & \textcolor{blue}{0.7416} & 0.6454 & 0.6463 & 0.6460 & 0.6438 & 0.6580 & 0.6531 
						& 0.6817 & \textcolor{blue}{0.6818} & \textcolor{red}{0.6869}&{0.6960}\\

	& LPIPS$\downarrow$ 	& \textcolor{blue}{0.3575} & \textcolor{red}{0.3546} & 0.1777 & 0.1615 & 0.1710 & 0.1750 & 0.1833 & 0.1613 
						& 0.1572&\textcolor{blue}{0.1530} & \textcolor{red}{0.1500}&{0.1388}\\

	& DISTS$\downarrow$ 	& \textcolor{blue}{0.2005} & \textcolor{red}{0.1996} & 0.1288 & 0.1158 & 0.1224 & 0.1252 & 0.1372 & 0.1165 
						& 0.1160& \textcolor{blue}{0.1139} & \textcolor{red}{0.1124}&{0.1104}\\

	& {LR-PSNR}$\uparrow$ 	& \textcolor{red}{\underline{52.52}} & \textcolor{blue}{\underline{52.35}} & 39.32 & 41.35 & 40.92 & 41.33 & \textcolor{red}{\underline{49.86}} & 40.99 
						& \textcolor{blue}{\underline{49.24}}& \underline{48.75} & \underline{49.19}& \underline{49.35} \\
\midrule
\multirow{5}*{General100}
	& PSNR$\uparrow$ 		& \textcolor{red}{30.30} & \textcolor{blue}{30.08} & 27.54 & 27.53 & 27.04 & 27.31 & 27.83 & 27.65 
						& 28.44 & \textcolor{blue}{28.57} & \textcolor{red}{28.74} & {29.12}\\

	& SSIM$\uparrow$ 		& \textcolor{red}{0.8696} & \textcolor{blue}{0.8662} & 0.7998 & 0.7984 & 0.7861 & 0.7899 & 0.7951 & 0.7995 
						& 0.8229& \textcolor{blue}{0.8250} & \textcolor{red}{0.8297}&{0.8400}\\

	& LPIPS$\downarrow$ 	& \textcolor{blue}{0.1665} & \textcolor{red}{0.1658} & 0.0961 & 0.0880 & 0.1084 & 0.0960 & 0.0962 & 0.0866 
						& 0.0784& \textcolor{blue}{0.0764} & \textcolor{red}{0.0753}&{0.0682}\\

	& DISTS$\downarrow$ 	& \textcolor{blue}{0.1321} & \textcolor{red}{0.1311} & 0.0955 & 0.0845 & 0.1166 & 0.0938 & 0.1022 & 0.0857 
						& 0.0831& \textcolor{blue}{0.0811} & \textcolor{red}{0.0795}&{0.0783}\\

	& {LR-PSNR}$\uparrow$ 	& \textcolor{red}{\underline{53.94}} & \textcolor{blue}{\underline{52.79}} & 41.44 & 41.93 & 40.05 & 41.84 & \underline{49.59} & 42.30 
						& \underline{49.82}& \textcolor{blue}{\underline{49.90}} & \textcolor{red}{\underline{50.11}}& {\underline{50.57}} \\
\midrule
\multirow{5}*{Urban100}
	& PSNR$\uparrow$ 		& \textcolor{red}{25.48} & \textcolor{blue}{25.21} & 22.84	 & 22.78	 & 22.74	 & 22.93	 & 23.68	 & 23.24 
						& 24.08 & \textcolor{blue}{24.21} 	 & \textcolor{red}{24.33} &{24.53}\\

	& SSIM$\uparrow$ 		& \textcolor{red}{0.8097} & \textcolor{blue}{0.8020} & 0.7196	 & 0.7214	 & 0.7107	 & 0.7169	 & 0.7316	 & 0.7365 
						& 0.7641 & \textcolor{blue}{0.7680} & \textcolor{red}{0.7707}  &{0.7784} \\

	& LPIPS$\downarrow$ 	& \textcolor{blue}{0.1960}	& \textcolor{red}{0.1961} & 0.1426	 & 0.1230	 & 0.1343	 & 0.1385	 & 0.1272	 & 0.1190 
						& 0.1090   	& \textcolor{blue}{0.1066} & \textcolor{red}{0.1065}  &{0.0988}\\

	& DISTS$\downarrow$ 	& \textcolor{blue}{0.1417}	 & \textcolor{red}{0.1409} & 0.1001	 & 0.0818	 & 0.0974	 & 0.0987	 & 0.0978	 & 0.0798 
						& 0.0783 	& \textcolor{blue}{0.0773}   & \textcolor{red}{0.0764} &{0.0774}\\

	& {LR-PSNR}$\uparrow$ 	& \textcolor{red}{\underline{51.21}} & \textcolor{blue}{\underline{50.52}} & 38.84	 & 39.70	 & 39.39	 & 39.07	 & \textcolor{red}{\underline{49.60}}	 & 40.40 
						& \underline{48.27}  & \underline{48.29}  & \textcolor{blue}{\underline{48.32}} & {\underline{48.52}} \\
\midrule
\multirow{5}*{Manga109} 
	& PSNR$\uparrow$ 		& \textcolor{red}{29.74} & \textcolor{blue}{29.36} & 26.26 & 26.50 & 26.07 & 26.04 & 27.11 & 26.74
						& 27.64& \textcolor{blue}{27.85}  & \textcolor{red}{28.08} &{28.61}\\

	& SSIM$\uparrow$ 		& \textcolor{red}{0.8997} & \textcolor{blue}{0.8948} & 0.8285 & 0.8245 & 0.8182 & 0.8117 & 0.8244 & 0.8267
						& 0.8440  & \textcolor{blue}{0.8493} & \textcolor{red}{0.8554} &{0.8737}\\

	& LPIPS$\downarrow$ 	& \textcolor{blue}{0.0975} & \textcolor{red}{0.0972} & 0.0709 & 0.0654 & 0.0716 & 0.0773 & 0.0663 & 0.0683
						& 0.0580 & \textcolor{blue}{0.0566}  & \textcolor{red}{0.0524} &{0.0431}\\

	& DISTS$\downarrow$ 	& \textcolor{blue}{0.0643} & \textcolor{red}{0.0605} & 0.0461 & 0.0397 & 0.0496 & 0.0488 & 0.0501 & 0.0403
						& 0.0407 & \textcolor{blue}{0.0382}  & \textcolor{red}{0.0351} &{0.0344} \\

	& {LR-PSNR}$\uparrow$ 	& \textcolor{red}{\underline{51.73}} & \textcolor{blue}{\underline{50.39}} & 40.35 & 40.68 & 38.96 & 39.83 & \underline{48.36} & 41.51
						& \underline{48.19} & \textcolor{blue}{\underline{48.49}} & \textcolor{red}{\underline{48.77}} & {\underline{49.33}} \\
\midrule
\multirow{5}*{DIV2K} 
	& PSNR$\uparrow$ 		& \textcolor{red}{29.48} & \textcolor{blue}{29.33} & 26.63 & 26.64 & 26.56 & 26.51 & 27.08 & 26.71 
						& 27.51 & \textcolor{blue}{27.57} & \textcolor{red}{27.69}  &{28.03}\\

	& SSIM$\uparrow$ 		& \textcolor{red}{0.8444} & \textcolor{blue}{0.8413} & 0.7625 & 0.7640 & 0.7578 & 0.7526 & 0.7558 & 0.7614 
						& 0.7890 & \textcolor{blue}{0.7906} & \textcolor{red}{0.7932}  &{0.8031}\\

	& LPIPS$\downarrow$ 	& \textcolor{blue}{0.2537} & \textcolor{red}{0.2530} & 0.1263 & 0.1154 & 0.1449 & 0.1217 & 0.1201 & 0.1100 
						& 0.1028 & \textcolor{blue}{0.0971} & \textcolor{red}{0.0957} &{0.0888}\\

	& DISTS$\downarrow$ 	& \textcolor{blue}{0.1261} & \textcolor{red}{0.1254} & 0.0613 & 0.0530 & 0.0858 & 0.0589 & 0.0622 & 0.0494 
						& 0.0513 & \textcolor{blue}{0.0492} & \textcolor{red}{0.0491}  &{0.0480}\\

	& {LR-PSNR}$\uparrow$ 	& \textcolor{red}{\underline{53.71}} & \textcolor{blue}{\underline{53.59}} & 40.87 & 42.61 & 40.40 & 41.90 & \underline{49.96} & 42.57 
						& \textcolor{blue}{\underline{50.54}} & \underline{50.36} & \textcolor{red}{\underline{50.80}}  & {\underline{51.04}} \\
\bottomrule
\end{tabular}
\end{center}
\label{tab:SROOE-quantitative-comparison}
\vspace{-0.4cm}
\end{table*}

\begin{figure*}[t!]
\setlength{\arrayrulewidth}{1.0pt}
\newcolumntype{Z}
{>{\centering\arraybackslash}X}
\renewcommand{\tabcolsep}{1pt}
\centering
\scriptsize
	\begin{tabularx}{\linewidth}{Z Z Z Z Z Z }
	{\includegraphics[width=0.9\linewidth]{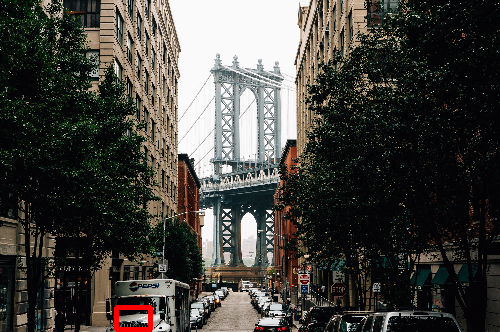}}&
	{\includegraphics[width=0.97\linewidth]{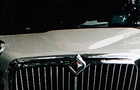}}&
	{\includegraphics[width=0.97\linewidth]{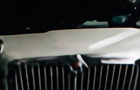}}&
	{\includegraphics[width=0.97\linewidth]{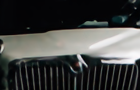}}&
	{\includegraphics[width=0.97\linewidth]{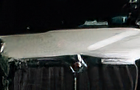}}&
	{\includegraphics[width=0.97\linewidth]{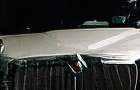}}\\
	0861 (DIV2K) &HR&RRDB&SROOE ($\mathbf{T}=\mathbf{0}$)&SRGAN&ESRGAN		\\
	&(PSNR$\uparrow$ / SSIM$\uparrow$ / LPIPS$\downarrow$)& (23.56 / 0.7929 / 0.2319) & (23.44 / 0.7894 / 0.2276) & (20.94 / 0.6912 / 0.1753) & (20.67 / 0.6827 / 0.1683)  \\
	{\includegraphics[width=0.9\linewidth]{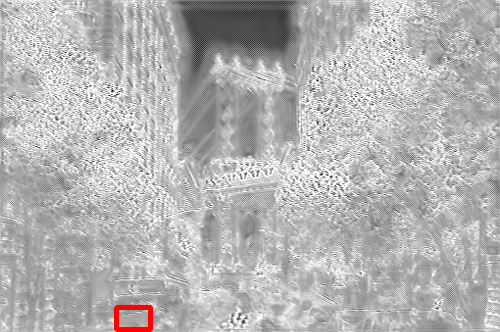}}&
	{\includegraphics[width=0.97\linewidth]{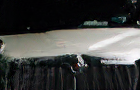}}	&
	{\includegraphics[width=0.97\linewidth]{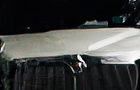}}&
	{\includegraphics[width=0.97\linewidth]{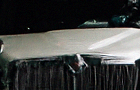}}	&
	{\includegraphics[width=0.97\linewidth]{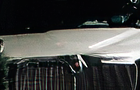}}	&
	{\includegraphics[width=0.97\linewidth]{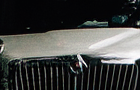}}	\\
	OOE Map ($\hat{\mathbf{T}}_{B})$ &SFTGAN&RankSRGAN&SRFlow&SPSR&SROOE ($\hat{\mathbf{T}}_{B}$)	\\
	& (21.05 / 0.6502 / 0.2018) & (21.28 / 0.6969 / 0.1696) & (21.69 / 0.7028 / 0.1619) & (20.61 / 0.6960 / 0.1602) & (\textcolor{red}{21.72} / \textcolor{red}{0.7375} / \textcolor{red}{0.1339}) \\
	\\[0.01em]
	{\includegraphics[width=0.90\linewidth]{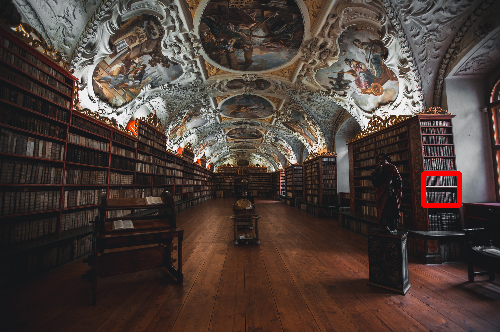}}&
	{\includegraphics[width=0.90\linewidth]{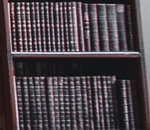}}&
	{\includegraphics[width=0.90\linewidth]{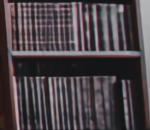}}&
	{\includegraphics[width=0.90\linewidth]{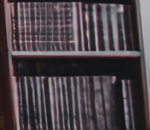}}&
	{\includegraphics[width=0.90\linewidth]{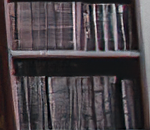}}&
	{\includegraphics[width=0.90\linewidth]{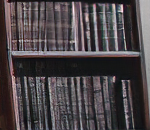}}\\
	0841 (DIV2K) &HR&RRDB&SROOE ($\mathbf{T}=\mathbf{0}$)&SRGAN&ESRGAN		\\
	&(PSNR$\uparrow$ / SSIM$\uparrow$ / LPIPS$\downarrow$)& (28.73 / 0.8851 / 0.1929)& (28.60 / 0.8813 / 0.1927)&(25.89 / 0.8128 / 0.1141) & (25.98 / 0.8182 / 0.1048)  \\
	{\includegraphics[width=0.90\linewidth]{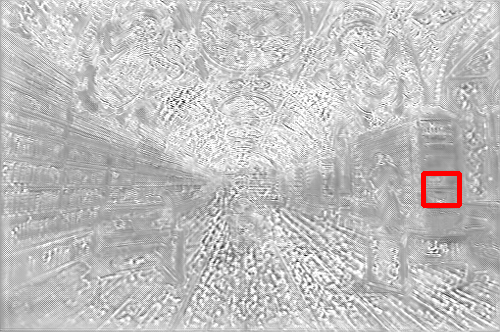}}&
	{\includegraphics[width=0.90\linewidth]{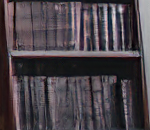}}&
	{\includegraphics[width=0.90\linewidth]{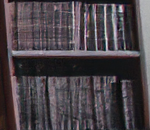}}&
	{\includegraphics[width=0.90\linewidth]{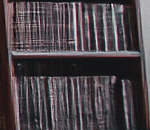}}	&
	{\includegraphics[width=0.90\linewidth]{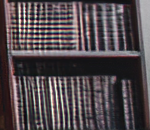}}	&
	{\includegraphics[width=0.90\linewidth]{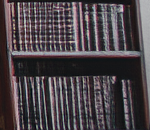}}	\\
	OOE Map ($\hat{\mathbf{T}}_{B}$)&SFTGAN&RankSRGAN&SRFlow&SPSR&SROOE \\
	& (25.76 / 0.8004 / 0.1370) & (25.83 / 0.8046 / 0.1098) & (26.45 / 0.7963 / 0.1093) & (26.32 / 0.8182 / 0.1076) & (\textcolor{red}{26.58} / \textcolor{red}{0.8283} / \textcolor{red}{0.0897}) \\
	\end{tabularx}
\vspace{-0.2cm}
\caption{Visual comparison with state-of-the-art SR methods. 
Among the seven perception-oriented SR methods, the best performances are highlighted in \textcolor{red}{red}.}
\label{fig:visual-comparison-SROOE-00}
\vspace{-0.9cm}
\end{figure*}

Although $\mathbf{T}^\ast_{S}$ is useful for training ${C}_{\psi}$, this pixel-wise objective selection without considering the interference caused by the convolutions of ${G}_{\theta}$ is not accurate ground truth.
Therefore, ${C}_{\psi}$ is optimized with three loss terms: pixel-wise objective map loss, pixel-wise reconstruction loss 
and perceptual loss, which measures the difference between the reconstructed and HR images, as follows:
\begin{equation}
{\psi}^\ast=\arg\min_{\psi}\mathbb{E}_{\mathcal{Z_{\mathbf{T}}}\sim\mathcal{P}_{\mathcal{Z_{\mathbf{T}}}}}\mathcal{L},
\label{eqn:opt_psi}
\end{equation}
\begin{equation}
\mathcal{L}=\lambda_{\mathbf{T}}\cdot\mathcal{L}_{\mathbf{T}}+{\lambda}^{OOE}_{rec}\cdot\mathcal{L}_{rec}+\lambda_{R}\cdot\mathcal{L}_{R},
\end{equation}
\begin{equation}
\mathcal{L}_{R}=\mathbb{E}\left[LPIPS\left(y,\hat{y}_{\hat{\mathbf{T}}_B}\right)\right],
\end{equation}
where $\mathcal{L}_{\mathbf{T}}$ and $\mathcal{L}_{rec}$ is the L1 losses between $\mathbf{T}^\ast_{S}$ and $\hat{\mathbf{T}}_{B}$ and between $y$ and $\hat{y}_{\hat{\mathbf{T}}_B}$, respectively. Meanwhile, $\mathcal{Z_{\mathbf{T}}}=(x,y,\mathbf{T}^\ast_{S})$ is the training dataset, and $\lambda_{\mathbf{T}}$, ${\lambda}^{OOE}_{rec}$ and $\lambda_{R}$ are the weights for each of the loss terms, respectively. 
During the ${C}_{\psi}$ model training, ${C}_{\psi}$ is combined with the already trained generative model, and the generator parameters are fixed. Therefore, losses for $C_\psi$ training, including LPIPS, are involved only in estimating locally-appropriate objective maps without affecting or changing the parameters of the generator. 

The architecture of $C_\psi$ consists of two separate sub-network: one is a feature extractor (F.E.) utilizing the VGG-19~\cite{SimonyanZ14a} and the other is a predictor with the UNet architecture~\cite{Olaf2015UNet}, as shown in Fig~\ref{fig:network-architecture}. For better performance, the feature extractor aims to get low to high-level features and delivers them to Unet, which makes the prediction. 
Since the structure of UNet has a wider receptive field, it is advantageous for predicting objectives in context. 

\section{Experiments}
\subsection{Experiment Setup}
\textbf{Materials and Evaluation Metrics.}
We use either the DIV2K~\cite{agustsson2017ntire} (800 images) or the DF2K \cite{lim2017enhanced} (3450 images) dataset to train our models.
Our test datasets include BSD100~\cite{Martin2001ICCV-BSD100}, General100~\cite{dong2016accelerating}, Urban100~\cite{huang2015single}, Manga109~\cite{matsui2017sketch}, and DIV2K validation set~\cite{agustsson2017ntire}. To evaluate the perceptual quality, we report LPIPS~\cite{zhang2018unreasonable} and DISTS~\cite{ding2020image}, which are full-reference metrics. DISTS is a perceptual metric that focuses on detail similarity. PSNR and SSIM~\cite{wang2004image} are also reported as fidelity-oriented metrics. The LR-PSNR metric is the PSNR between the LR input and downscaled SR images. The higher the LR-PSNR, the better the consistency between the SR results and LR images, where 45 dB or more is recommended for good LR consistency as addressed in NTIRE challenge~\cite{Lugmayr_2021_CVPR}. Because consistency with the LR input images is important, we also report the LR-PSNR.

\vspace{0.4cm}
\textbf{Training Details.}
The HR and LR batch sizes are ${256}\times{256}$ and ${64}\times{64}$, respectively. All training parameters are set to be equal to those of ESRGAN~\cite{wang2018esrgan}, except for the loss weights. For the generator training, $t$ is a random variable with uniform distribution in [0, 1]. 
$\boldsymbol{\alpha}$=$[{1}\times{10}^{-2},1,\boldsymbol{1}]$ and $\boldsymbol{\beta}$=$[{1}\times{10}^{-2},0,\boldsymbol{0}]$.

\newcolumntype{M}[1]{>{\centering\arraybackslash}m{#1}}
\begin{table}[!t]
\caption{Comparison of performance according to different selections. The bold checkmark indicates the change from the left selection. The 1st and the 2nd best performances except for SROOS are highlighted in
 \textcolor{red}{red} and \textcolor{blue}{blue}, respectively.}
\vspace{-0.5cm}
\begin{center}
\scriptsize
\renewcommand{\tabcolsep}{1pt}
\begin{tabular}{M{11mm}|M{15mm}|M{11mm}|M{9mm}M{9mm}M{9mm}|M{14mm}}
\toprule
	\multicolumn{2}{c|}{Methods}	  		 						&  SROT & \multicolumn{3}{c|}{SROOE}  	 & SROOS \\ 
\midrule
	{Objective}  				& P2  		 					& $\checkmark$  	& $\checkmark$  	&   			&			& \\ 
	Trajectory							& P1234  	 						&			& 			& $\boldcheckmark$	&$\checkmark$	& $\checkmark$	\\ 
\midrule
	\multirow{3}*{$\mathbf{T}$}  		& $\mathbf{T}_{t=0.8}$  	&$\checkmark$ &			& 			&			& \\ 
								& $\hat{\mathbf{T}}_{B}$ (OOE) 		&  			& $\boldcheckmark$	& $\checkmark$	&$\checkmark$	& \\ 
								& $\mathbf{T}^\ast_{S}$ (OOS) 		&  			& 			& 			&			& $\boldcheckmark$ 	\\
\midrule 
			{training}  			& DIV2K  							&$\checkmark$	&$\checkmark$	&$\checkmark$	&    			& \\ 
			{DB} 			& DF2K 							&  			& 			& 			&$\boldcheckmark$	& $\checkmark$ \\ 
\midrule

\multirow{4}*{Metric}
	& PSNR$\uparrow$ 		& 26.49 & 26.53 & \textcolor{blue}{26.65} & \textcolor{red}{26.74} & {27.07} \\

	& SSIM$\uparrow$ 		& 0.7803 & 0.7816 & \textcolor{blue}{0.7853} & \textcolor{red}{0.7872} &{0.7982} \\

	& LPIPS$\downarrow$ 	& 0.1011 & 0.1000 & \textcolor{blue}{0.0978} & \textcolor{red}{0.0960} &{0.0875} \\

	& LR-PSNR$\uparrow$ 	& 49.21 & 49.21 & \textcolor{blue}{49.32} & \textcolor{red}{49.44} &{49.76} \\
\bottomrule
\end{tabular}
\end{center}
\label{tab:ablation-SROOE-01}
\vspace{-0.7cm}
\end{table}

\subsection{Evaluation}
\textbf{Quantitative Comparison.}
Table~\ref{tab:SROOE-quantitative-comparison} shows the quantitative performance comparison for the $4\times$ SR. We compared it with a distortion-oriented method, RRDB~\cite{wang2018esrgan}, and perception-oriented methods, such as SRGAN~\cite{ledig2017photo}, ESRGAN~\cite{wang2018esrgan}, SFTGAN~\cite{wang2018recovering}, RankSRGAN~\cite{zhang2019ranksrgan}, SRFlow~\cite{lugmayr2020srflow}, SPSR~\cite{ma2020structure}, and FxSR~\cite{Park2022Access-FxSR}. The table shows that our method yields the best results among the perception-oriented methods on all datasets, not only in terms of LPIPS~\cite{zhang2018unreasonable} and DISTS~\cite{ding2020image}, but also in terms of distortion-oriented metrics such as PSNR and SSIM. It also exceeds 45 dB in LR-PSNR, indicating that LR consistency is well-maintained, as addressed in NTIRE~\cite{Lugmayr_2021_CVPR}.
In addition, SROOE using a local objective map outperforms SROT with the globally optimal $t$ value for the Urban100 and DIV2K benchmarks in terms of both LPIPS and PSNR in Fig.~\ref{fig:PD-curve-t}. SROOS with $\mathbf{T}^\ast_{S}$ has the best PSNR, SSIM, LPIPS, and DISTS scores, which shows the approximated upper bounds of the proposed SROOE. On the other hand, when the objective map $\mathbf{T}$ is set to be $\mathbf{0}$, SROOE operates as a distortion-oriented SR model. Although it is slightly inferior to RRDB~\cite{wang2018esrgan} in terms of PSNR, its performance is not far behind while showing better LPIPS. This implies that SROOE performs close to RRDB~\cite{wang2018esrgan} for the regions needing distortion-oriented restoration, and thus the overall distortion is reduced while achieving high perceptual quality.

\textbf{Qualitative Comparison.}
Fig. \ref{fig:visual-comparison-SROOE-00} shows a visual comparison, where we can observe that SROOE generates more accurate structures and details. In particular, it appears that there is little change in the structural component between the SROOE results using $\mathbf{T}=\mathbf{0}$ and $\hat{\mathbf{T}}_{B}$, and sharp edges and generated details are added to the structural components. 
Additional visual comparisons are provided in the supplementary.

\section{Ablation Study}

Table \ref{tab:ablation-SROOE-01} reports average values in terms of each metric on all five benchmarks in Table~\ref{tab:SROOE-quantitative-comparison}, which vary according to the change in each element.
The two different objective trajectories shown in Fig.~\ref{fig:vector-function}(b) and (d) are referred to as P1234 and P2, respectively.  The table shows that the SR performance improves step by step, when going from P2, fixed $t$, and DIV2K  to P1234, OOE, and DF2K, respectively.
Specifically, our proposed model, SROOE-P1234 trained with DF2K is improved by 0.25 dB in PSNR, 0.0069 in SSIM, 0.0051 in LPIPS, and 0.23 dB in LR-PSNR compared to SROT-P2 corresponding to FxSR with $t$=0.8~\cite{Park2022Access-FxSR}. 
Comparing the performance of SROOE and SROOS shows that there is still room for performance improvement.

\newcolumntype{M}[1]{>{\centering\arraybackslash}m{#1}}
\begin{table}[!t]
\caption{Comparison of the running time and the SR model size.}
\vspace{-0.5cm}
\begin{center}
\scriptsize
\renewcommand{\tabcolsep}{1pt}
\begin{tabular}{M{18mm}M{15mm}M{15mm}M{15mm}M{15mm}}
\toprule
& SRGAN\cite{ledig2017photo}& ESRGAN\cite{wang2018esrgan}& FxSR~\cite{Park2022Access-FxSR} & SROOE\\
\midrule
Run Time (msec) & 0.014 & 0.138 & 0.501 & 0.968 \\
Param Size (MB) & 1.51 & 16.69 & 18.30 & 70.20 \\
\bottomrule
\end{tabular}
\end{center}
\label{tab:ablation-complexity}
\vspace{-0.7cm}
\end{table}

\textbf{Complexity Analysis.}
A comparison of the running times and parameter sizes is presented in Table~\ref{tab:ablation-complexity}. The complexity of the $4\times$ SR of one ${128}\times{128}$ LR image is measured using an NVIDIA RTX3090 GPU. 

\textbf{Limitations.}
Although applying locally appropriate objectives can significantly improve the LR to HR mapping accuracy, even if the generator uses an optimal objective map $\mathbf{T}^\ast_{S}$, it is still limited in achieving full reconstruction. 
This means that the proposed generator is still unable to generate all HRs with the objective set used for training in this study, 
and thus, more sophisticated perceptual loss terms than the VGG feature space are still required to overcome this.
And still, there remains a limit to solving the ill-posed problem caused by high-frequency loss.

\section{Conclusion}
We proposed a novel SISR framework for perceptually accurate HR restoration, where an objective estimator provides an optimal combination of objectives for a given image patch, and a generator produces SR results that reflect the target objectives.
For this purpose, we employed objective trajectory learning to efficiently train a single generative model that can apply varying objective sets. Experiments show that the proposed method reduces visual artifacts, such as structural distortion and unnatural details, and achieves improved results compared to those of state-of-the-art perception-oriented methods in terms of both perceptual and distortion metrics. The proposed method can be applied to off-the-shelf and other SISR network architectures.


{\small
\bibliographystyle{ieee_fullname}
\bibliography{egbib_my}

}

\end{document}